\documentclass[10pt,logo,copyright]{engineai-report}
\usepackage{iftex}
\ifPDFTeX
    \usepackage[utf8]{inputenc} %
    \usepackage[T1]{fontenc}    %
\fi

\usepackage{parskip}        %
\usepackage{url}            %
\usepackage{booktabs}       %
\usepackage{amsfonts}       %
\usepackage{nicefrac}       %
\usepackage{microtype}      %
\usepackage[dvipsnames]{xcolor} %
\usepackage{graphicx}
\usepackage{animate}        %
\usepackage{subcaption}
\usepackage{tabularx}
\usepackage{makecell}
\usepackage{adjustbox}
\usepackage{setspace}
\usepackage{todonotes}
\usepackage{colortbl}
\usepackage{wrapfig}

\definecolor{rankbest}{RGB}{144,238,144}
\definecolor{rankmid}{RGB}{255,255,180}
\definecolor{rankworst}{RGB}{248,150,150}
\definecolor{deltadarkpos}{RGB}{50,180,50}
\definecolor{deltalightpos}{RGB}{180,235,180}
\definecolor{deltayellow}{RGB}{255,255,200}
\definecolor{deltalightneg}{RGB}{250,180,180}
\definecolor{deltadarkneg}{RGB}{220,80,80}

\newcolumntype{M}[1]{>{\centering\arraybackslash}m{#1}}
\usepackage{float}
\usepackage{tikz}
\usetikzlibrary{positioning,shapes,arrows,arrows.meta,fit,tikzmark}
\usepackage{amsmath,amsfonts,bm, bbm,leftindex}
\usepackage{multirow}
\usepackage{comment}
\usepackage{lipsum}
\usepackage[para]{threeparttable}

\def\eqref#1{equation~\ref{#1}}

\def\1{\bm{1}}

\DeclareMathAlphabet{\mathsfit}{\encodingdefault}{\sfdefault}{m}{sl}
\SetMathAlphabet{\mathsfit}{bold}{\encodingdefault}{\sfdefault}{bx}{n}

\makeatletter
\let\save@mathaccent\mathaccent
\newcommand*\if@single[3]{%
  \setbox0\hbox{${\mathaccent"0362{#1}}^H$}%
  \setbox2\hbox{${\mathaccent"0362{\kern0pt#1}}^H$}%
  \ifdim\ht0=\ht2 #3\else #2\fi
  }
\newcommand*\rel@kern[1]{\kern#1\dimexpr\macc@kerna}
\newcommand*\widebar[1]{\@ifnextchar^{{\wide@bar{#1}{0}}}{\wide@bar{#1}{1}}}
\newcommand*\wide@bar[2]{\if@single{#1}{\wide@bar@{#1}{#2}{1}}{\wide@bar@{#1}{#2}{2}}}
\newcommand*\wide@bar@[3]{%
  \begingroup
  \def\mathaccent##1##2{%
    \let\mathaccent\save@mathaccent
    \if#32 \let\macc@nucleus\first@char \fi
    \setbox\z@\hbox{$\macc@style{\macc@nucleus}_{}$}%
    \setbox\tw@\hbox{$\macc@style{\macc@nucleus}{}_{}$}%
    \dimen@\wd\tw@
    \advance\dimen@-\wd\z@
    \divide\dimen@ 3
    \@tempdima\wd\tw@
    \advance\@tempdima-\scriptspace
    \divide\@tempdima 10
    \advance\dimen@-\@tempdima
    \ifdim\dimen@>\z@ \dimen@0pt\fi
    \rel@kern{0.6}\kern-\dimen@
    \if#31
      \overline{\rel@kern{-0.6}\kern\dimen@\macc@nucleus\rel@kern{0.4}\kern\dimen@}%
      \advance\dimen@0.4\dimexpr\macc@kerna
      \let\final@kern#2%
      \ifdim\dimen@<\z@ \let\final@kern1\fi
      \if\final@kern1 \kern-\dimen@\fi
    \else
      \overline{\rel@kern{-0.6}\kern\dimen@#1}%
    \fi
  }%
  \macc@depth\@ne
  \let\math@bgroup\@empty \let\math@egroup\macc@set@skewchar
  \mathsurround\z@ \frozen@everymath{\mathgroup\macc@group\relax}%
  \macc@set@skewchar\relax
  \let\mathaccentV\macc@nested@a
  \if#31
    \macc@nested@a\relax111{#1}%
  \else
    \def\gobble@till@marker##1\endmarker{}%
    \futurelet\first@char\gobble@till@marker#1\endmarker
    \ifcat\noexpand\first@char A\else
      \def\first@char{}%
    \fi
    \macc@nested@a\relax111{\first@char}%
  \fi
  \endgroup
}
\makeatother

\definecolor{darkred}{rgb}{0.7, 0.0, 0.0}

\usepackage{amsmath} 
\usepackage{marvosym}
\usepackage[normalem]{ulem}
\usepackage{multirow}

\usepackage{pifont}
\usepackage{graphicx}

\newcommand{\ourMthd}{Awaken-HI}
\newcommand{\ourData}{Awaken-Human}

\usepackage[nameinlink]{cleveref}
\crefname{equation}{Eq.}{Eqs.}
\crefname{figure}{Fig.}{Figs.}
\crefname{section}{Sec.}{Sec.}
\crefname{appendix}{App.}{App.}
\crefname{table}{Tab.}{Tabs.}
\crefname{algorithm}{Algo}{Algo}
\crefname{thm}{Thm}{Thm}
\Crefname{thm}{Thm}{Thm}
\crefname{prop}{Prop}{Prop}

\newcommand{\crefnames}[3]{%
  \@for\next:=#1\do{%
    \expandafter\crefname\expandafter{\next}{#2}{#3}%
  }%
}

\title{\ourMthd{}: Scaling Humanoid Intelligence with Whole-Body Human Experience}

\author{%
\centerline{\textbf{\reportorganization}}
\centerline{\normalfont\small Project Page: \href{\reportwebsite}{\texttt{\reportwebsite}}}
}
\usepackage[numbers,sort&compress]{natbib}
\setcitestyle{numbers}
\begin{document}
\maketitle

\begin{figure}[H]
    \centering
    \captionsetup{type=figure, justification=justified, singlelinecheck=false}
    \includegraphics[width=\linewidth]{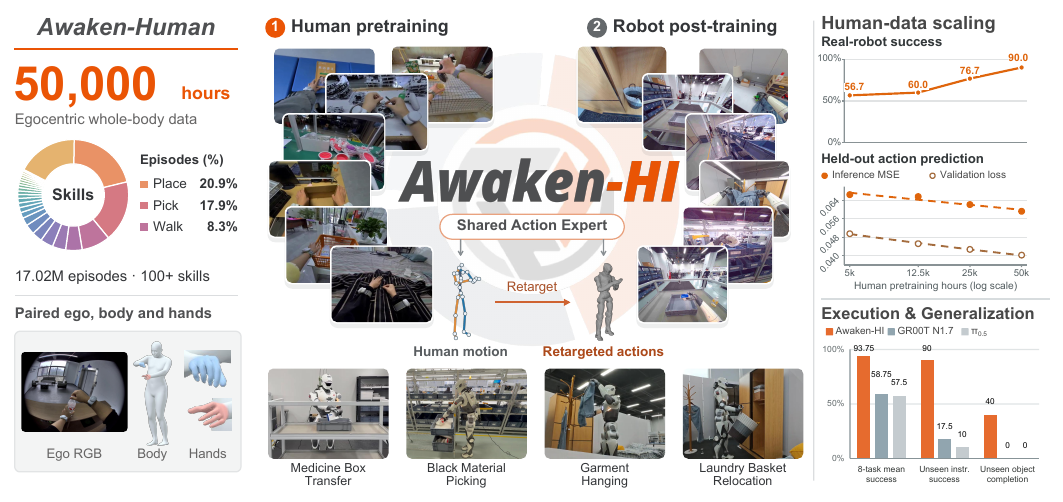}
    \caption{\textbf{Overview of \ourMthd{} and the \ourData{} dataset.} \ourData{} pairs approximately 50,000 hours of egocentric stereo video with synchronized whole-body motion (left). We pretrain \ourMthd{} on human motion and retargeted robot actions, then fine-tune it on robot demonstrations (center). The right panels summarize scaling and generalization results alongside the full model's 93.75\% mean success rate across eight individual tasks.}
    \label{fig:teaser}
\end{figure}

\begin{abstract}

A central goal of humanoid robotics is for robots to perform the everyday work people do in homes and workplaces.
Existing egocentric video datasets cover many of these activities but provide little synchronized whole-body motion data for learning loco-manipulation at scale.
We build \ourData{}, which pairs approximately 50,000 hours of egocentric stereo video with synchronized whole-body human motion and covers more than 100 skills.
Building on this, we also introduce \ourMthd{} (Humanoid Intelligence), a vision-language-action model for humanoid loco-manipulation that jointly predicts original human motion and retargeted robot actions through a shared action expert.
We show that loco-manipulation skills transfer more effectively from people to humanoids when human trajectories directly supervise the shared action expert.
Increasing human pretraining data reduces held-out robot action prediction error according to an empirical scaling law and raises the mean success rate on downstream real-robot tasks.
After full-corpus pretraining and T800 adaptation, \ourMthd{} achieves a mean success rate of 93.75\% across eight individual tasks and outperforms GR00T N1.7 and $\pi_{0.5}$ in both individual and long-horizon evaluations.
The same policy generalizes to objects unseen during robot post-training and to new pairings of familiar objects and destinations.

\end{abstract}

\abscontent

\section{Introduction}
\label{sec:intro}

Humanoid robots could help people with a wide range of tasks in homes and workplaces.
Many require whole-body loco-manipulation, in which the robot coordinates its steps and posture with its hand movements~\cite{jiang2026wholebodyvla,li2026omega}.
Vision-language-action (VLA) models~\cite{pi0,pi0.5,gr00t} and world action models~\cite{ye2026world,lingbotva,fastwam} have shown that robots can learn diverse manipulation skills from demonstrations.
For humanoids, however, the cost of whole-body teleoperation limits the range of tasks and environments represented in robot demonstrations~\cite{he2024omnih2o,ze2026twist2}.

Human experience of everyday activity offers a more scalable alternative to robot teleoperation.
Recent work draws on the variety of activities and environments in these videos~\cite{gigaworldpolicy,wu2024unleashing} to pretrain manipulation policies~\cite{wei2026psi0,zheng2026egoscale,li2026scalable}.
Despite this diversity, visual observations alone provide no explicit targets for whole-body motion.
Prior work therefore retargets recorded human motion to obtain robot action targets~\cite{chen2026egohumanoid,post2026a0}.
However, differences in body proportions and joint structure can cause the resulting postures and hand trajectories to deviate from those in the human demonstration~\cite{araujo2026retargeting}.
These deviations motivate retaining the original human motion as a training signal alongside the retargeted robot actions.
Using that signal at scale requires egocentric video paired with synchronized whole-body motion across diverse tasks and environments.
Yet many large egocentric datasets provide video alone or primarily track the head and hands~\cite{grauman2022ego4d,hoque2026egodex}, while datasets such as Nymeria~\cite{ma2024nymeria} offer only a few hundred hours of paired recordings.

We address this data gap with \ourData{}, a dataset for humanoid pretraining that pairs approximately 50,000 hours of egocentric stereo video with synchronized whole-body human motion.
Collected under a common protocol using lightweight wearable sensors, the demonstrations cover more than 100 skills across homes, workplaces, retail spaces, and storage areas.
We reconstruct 3D hand motion from the video and annotate the demonstrations with subtask descriptions and frame-level depth maps.
For each demonstration, we retain the original human trajectory alongside the corresponding retargeted robot actions.

Building on \ourData{}, we introduce \ourMthd{}, a VLA model for humanoid loco-manipulation.
By predicting human motion alongside retargeted robot actions, its shared action expert receives direct supervision from the original human trajectories.
During pretraining, the expert also predicts camera-frame wrist poses, while separate heads estimate current and future depth and describe the current subtask.
We then post-train the policy on approximately 150 hours of T800 teleoperation data using the same robot action interface.

We observe that \ourMthd{}'s robot action prediction error on held-out data follows an empirical scaling law, decreasing as the pretraining set grows from 5,000 to 50,000 hours of human demonstrations.
With the same T800 demonstrations for post-training, more human pretraining data also raises the policy's mean success rate on three household tasks.
Controlled ablations show that adding human motion supervision reduces action prediction error and improves real-robot task success compared with pretraining on retargeted actions alone.
After full-corpus pretraining, we post-train \ourMthd{} on the same T800 demonstrations used for GR00T N1.7~\cite{gr00t} and $\pi_{0.5}$~\cite{pi0.5}.
\ourMthd{} outperforms both baselines on all eight individual loco-manipulation tasks, averaging 93.75\% success.
It retains this advantage on long-horizon industrial and household tasks.
The policy also follows instructions that pair familiar objects and destinations in new ways and applies a learned manipulation skill to objects unseen during robot post-training.

Our main contributions are as follows:
\begin{itemize}
    \item We introduce \textbf{\ourMthd{}}, a VLA model for humanoid loco-manipulation that jointly predicts original human motion and robot actions through a shared action expert.
    \item We build \textbf{\ourData{}}, a dataset pairing 50,000 hours of egocentric stereo video with synchronized whole-body human motion across diverse tasks and environments.
    \item We find that as human pretraining data grows, action prediction error decreases according to an empirical scaling law and real-robot success rates rise. \ourMthd{} also generalizes to new pairings of familiar objects and destinations and to objects unseen during robot post-training.
\end{itemize}

\newcommand{\bkmark}{\textcolor{black}{\ding{51}}}
\begin{table}[H]
\centering
\footnotesize
\setlength{\tabcolsep}{5pt}
\renewcommand{\arraystretch}{1.15}
\caption{\textbf{Comparison with existing egocentric human datasets.}
Hours and episodes are those released by the original source, and -- denotes a value the source does not report.}
\label{tab:data_comparison}
\begin{tabular}{l r l l c c c}
\toprule
Dataset & Hours & Episodes & Capture setup & Whole body & Hands & Depth \\
\midrule
Egocentric-100K~\cite{buildaiegocentric100k2025}     & 100,405 & 2.01M & Head-mounted fisheye             &              &              &              \\
EgoScale~\cite{zheng2026egoscale}            & 20,854  & --    & Multiple human-data sources      &              & \bkmark      &              \\
Ego4D~\cite{grauman2022ego4d}               & 3,670   & --    & Head-worn cameras                &              &              &              \\
EgoVerse~\cite{punamiya2026egoverse}            & 1,362   & 80K   & Aria / custom hardware           &              & \bkmark      &              \\
Eidon Tracker POV~\cite{eidon2026trackerpov}   & 1,273.8 & 13.5K & Head camera and arm IMUs         &              &              &              \\
EgoDex~\cite{hoque2026egodex}              & 829     & 338K  & Apple Vision Pro                 &              & \bkmark      &              \\
HoloAssist~\cite{HoloAssist2023}          & 166     & 2.2K  & HoloLens 2                       &              & \bkmark      & \bkmark      \\
\midrule
EgoSuite-Open100K~\cite{lightwheel2026egosuite}   & 10,000$^{\ddagger}$ & -- & Head camera, body subsets & \bkmark   & \bkmark      &              \\
Ego-Exo4D~\cite{grauman2024ego}           & 1,286$^{\dagger}$ & -- & Aria and exocentric cameras & \bkmark       & \bkmark      & \bkmark      \\
HumanVerse~\cite{post2026a0}          & 500     & 33K   & PICO and trackers, GoPro         & \bkmark      & \bkmark      &              \\
Nymeria~\cite{ma2024nymeria}             & 300     & --    & Aria, wristbands, XSens          & \bkmark      &              &              \\
\midrule
\rowcolor{gray!18}
\textbf{\ourData{} (Ours)} & \textbf{50,000} & \textbf{17.0M} & PICO and trackers & \bkmark & \bkmark & \bkmark \\
\bottomrule
\end{tabular}

\vspace{2pt}
{\footnotesize
$^{\dagger}$Ego-Exo4D counts egocentric and exocentric recordings together.
$^{\ddagger}$EgoSuite-Open100K plans 100,000 hours, of which 10,000 are released; its full-body subsets are planned, and their share of the released hours is not documented.\par}
\end{table}

\section{Related Work}
\label{sec:related_work}

\subsection{Foundation Models for Robot Control}
\label{sec:related_foundation_models}

Generalist robot policies map observations and instructions to actions across a wide range of manipulation tasks~\cite{brohan2022rt,team2024octo,kim2024openvla,pi0,wu2026foundation,gr00t,pi0.5}.
For a humanoid, loco-manipulation additionally requires moving and staying balanced while adjusting posture to reach objects at different heights and distances.
One line of work separates the control interfaces for the upper and lower body.
WholeBodyVLA~\cite{jiang2026wholebodyvla} and $\Psi_0$~\cite{wei2026psi0} predict upper-body actions and coarse base commands, which a locomotion controller converts into lower-body motion.
More recent approaches predict whole-body motion, enabling the policy to learn coordination between locomotion and manipulation~\cite{zheng2026motionwam,li2026omega,post2026a0}.
Our \ourMthd{} follows this approach and learns motion priors through large-scale human pretraining.

\subsection{Egocentric Data and Motion Alignment}
\label{sec:related_motion_alignment}

Egocentric datasets have grown rapidly in scale, but largely at the cost of recorded whole-body motion or a consistent protocol.
Nymeria~\cite{ma2024nymeria} uses an inertial motion-capture suit, and HumanVerse in $\lambda_0$~\cite{post2026a0} uses a headset with body-worn trackers, yet each covers only a few hundred hours.
Larger datasets such as Ego4D~\cite{grauman2022ego4d}, EgoVerse~\cite{punamiya2026egoverse}, and EgoDex~\cite{hoque2026egodex} provide video with at most head, hand, or upper-body poses, while Ego-Exo4D~\cite{grauman2024ego} derives body pose from exocentric cameras.
Pretraining corpora such as EgoScale~\cite{zheng2026egoscale} and Ego4WAM~\cite{sun2026ego4wam} exceed 10,000 hours by aggregating many sources, which differ in sensors and annotation pipelines.
We build \ourData{} to fill this gap with 50,000 hours of egocentric video and whole-body motion recorded by wearable trackers, all captured under the same protocol.
Table~\ref{tab:data_comparison} compares \ourData{} with other egocentric datasets in terms of scale, capture setup, and annotation coverage.

To construct robot-action supervision, human motion is mapped to robot embodiments through direct joint mappings~\cite{fu2024humanplus} or geometric retargeting, as in GMR~\cite{araujo2026retargeting}.
Differences in body proportions and joint structure can alter the resulting postures and end-effector trajectories~\cite{araujo2026retargeting,he2024learning}.
Controller dynamics introduce further discrepancies between references and execution~\cite{chen2026egohumanoid,jiang2026egoalign}.
Controller-aware refinement~\cite{chen2026egohumanoid,zhao2026halomi} and causal state reconstruction, as in EgoAlign~\cite{jiang2026egoalign}, improve the consistency of human-derived supervision with robot behavior.
Our annotation pipeline replaces noisy hand annotations with 3D hand poses reconstructed from egocentric video and uses them to refine end-effector alignment, reducing end-effector errors relative to GMR.

\subsection{Learning from Egocentric Demonstrations}
\label{sec:related_egocentric_learning}

Human demonstrations are typically used in embodied learning to supervise perception, world modeling, and action learning.
R3M~\cite{nair2022r3m} pretrains a visual encoder on egocentric human video so that manipulation policies can be learned from fewer robot demonstrations.
World-action models instead use future frames as prediction targets, learning how scenes change as people interact with objects even without action labels~\cite{zheng2026motionwam,sun2026ego4wam}.
When hand or body trajectories are available, human demonstrations can also supervise action learning~\cite{qiu2025humanoid,yang2025egovla,wei2026psi0}.
EgoMimic~\cite{kareer2025egomimic} aligns human hand trajectories with robot end-effector trajectories in a common coordinate frame to train a shared manipulation policy.
EgoScale~\cite{zheng2026egoscale} and $\lambda_0$~\cite{post2026a0} retarget human motion into robot hand or whole-body actions and train on these labels.
However, constrained by morphological differences, retargeting inevitably introduces global errors and distorted postures~\cite{araujo2026retargeting,chen2026egohumanoid}.
We therefore train \ourMthd{} on the original human motion together with the retargeted robot actions, learning motion priors directly from the demonstrations.

\section{\ourData{}}
\label{sec:data}

\begin{figure}[t]
    \centering
    \captionsetup{type=figure, justification=justified, singlelinecheck=false}
    \includegraphics[width=\linewidth]{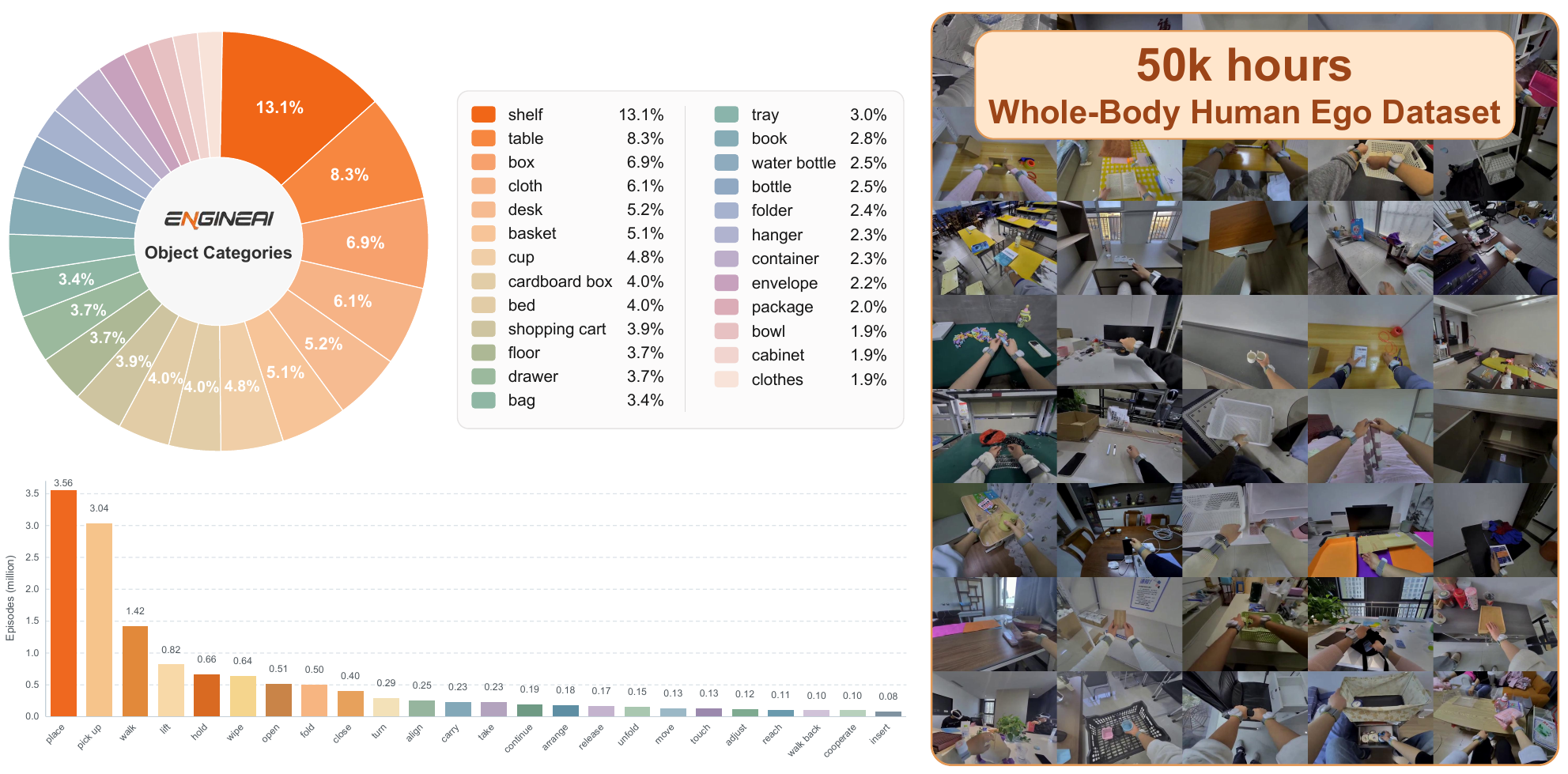}
    \caption{
        \textbf{Object and skill coverage in \ourData{}.} The dataset contains 17.02 million episodes spanning more than 100 skills. The left panels show the distribution of object categories (top) and episode counts for selected skills (bottom). The right panel shows representative egocentric frames.
    }
    \label{fig:data_overview}
\end{figure}

\subsection{Large-Scale Egocentric Corpus}
\label{sec:egocentric_corpus}

We build \ourData{} to study how the scale of human demonstrations affects humanoid pretraining. 
The corpus pairs 50,000 hours of egocentric video with synchronized whole-body human motion recorded by wearable sensors, making it the largest such dataset to our knowledge.
It comprises approximately 17.02 million episodes, 5.41 billion pose frames, and 11.18 million valid language instructions.

Our setup consists of a PICO headset and five body-worn trackers, synchronously recording stereo RGB video and skeletal sequences from PICO’s body-tracking system~\cite{chen2026egohumanoid,jiang2026egoalign}. Collectors follow task goals and adapt their walking paths, stance, and grasping strategies to the scene, preserving natural repositioning and motion transitions within continuous interactions. The corresponding task instructions are retained alongside raw video, motion trajectories, and camera and device parameters for subsequent annotation and alignment.

The demonstrations cover more than 100 skills across homes, workplaces, retail spaces, and storage areas. Activities range from picking up, carrying, and placing objects to folding garments and wiping surfaces. Continuous recordings capture how people coordinate walking, turning, and changes in posture with these hand movements. Figure~\ref{fig:data_overview} shows representative egocentric views, the distribution of interaction objects such as shelves, tables, boxes, and everyday containers, and episode counts for selected skills.

\begin{figure}[t]
    \centering
    \captionsetup{type=figure, justification=justified, singlelinecheck=false}
    \includegraphics[width=\linewidth]{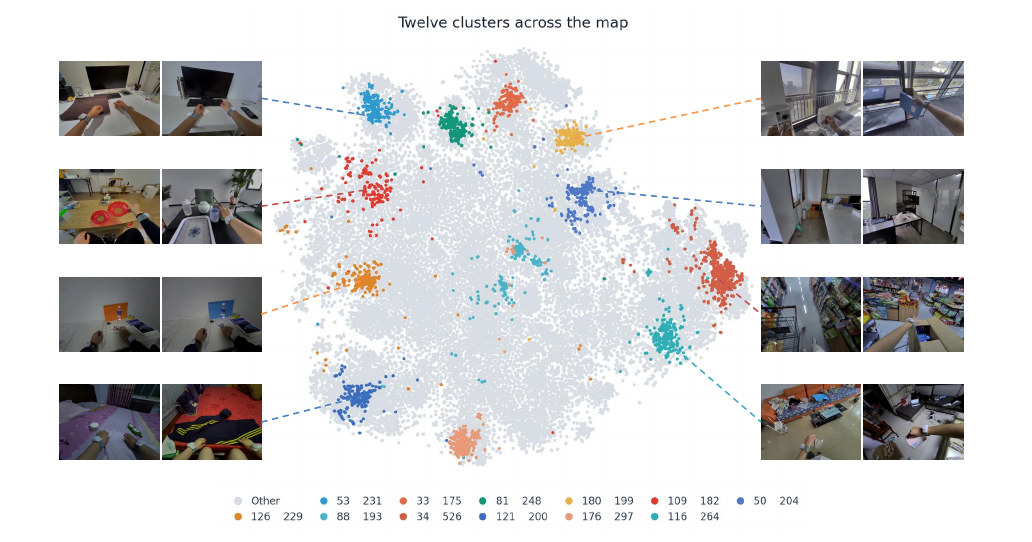}
    \caption{\textbf{Visual diversity in \ourData{}.} Each point represents a video sample in a two-dimensional t-SNE projection of V-JEPA~2 features. Selected clusters appear in color and the remaining samples in gray. Dashed lines link the example frames on either side to their corresponding clusters.
    }
    \label{fig:data_distribution}
\end{figure}

\subsection{Egocentric Data Annotation}
\label{sec:data_curation_alignment}

We develop an annotation pipeline for whole-body egocentric demonstrations that first undistorts raw stereo fisheye images using the camera calibration parameters. It then performs temporal segmentation and semantic annotation, visual hand-motion reconstruction, and stereo depth estimation to produce time-aligned language, human motion, and depth supervision for pretraining.

\noindent\textbf{Semantic annotation.} We fine-tuned a vision-language model (VLM) for egocentric demonstration annotation. The model produces initial annotations comprising temporal segment boundaries, segment-level language descriptions, and atomic action labels selected from a predefined library. Human reviewers verify and correct these initial annotations to produce segment-level semantic labels for training and data selection.

\noindent\textbf{Visual hand re-annotation.} The PICO hand annotations contain tracking gaps, positional errors, and temporal jitter that affect supervision of fine-grained hand motion. To address these issues, we use our in-house 3D hand reconstruction model to generate MANO-based 21-keypoint hand skeleton annotations from egocentric video~\cite{romero2022embodied}. The model generates hand-quality labels alongside the skeleton annotations, using hand visibility ratios to identify low-quality segments with incomplete or occluded hand observations during manipulation. We use these quality labels to exclude low-quality segments and replace the original PICO hand annotations with the retained reconstructions. These annotations provide human hand motion supervision and reference targets for subsequent robot end-effector alignment.

\noindent\textbf{Stereo depth annotation.} We generate dense depth annotations from the synchronized stereo observations collected with PICO. We apply Fast-FoundationStereo~\cite{wen2026fast} to undistorted, rectified image pairs to estimate disparity, then convert it to metric depth using the calibrated intrinsics and stereo baseline. The resulting depth maps are paired with their corresponding RGB frames and retained as geometric annotations of the demonstrations.

\subsection{Human-to-Humanoid Alignment and Dataset Construction}
\label{sec:teleoperation_data}

We align annotated human demonstrations with the T800 action space through whole-body retargeting and end-effector refinement. The resulting trajectories undergo quality filtering, corpus curation, and dataset partitioning, while the corresponding human motion annotations are retained for joint pretraining.

\noindent\textbf{Robot action target construction.}
We develop a T800-specific retargeting pipeline based on GMR~\cite{araujo2026retargeting}. We first scale the recorded human skeletons to T800's height and use visually re-annotated hand motion to align the end-effector references. We then retarget the aligned human motion to robot root and joint trajectories and optimize wrist orientations to refine end-effector alignment. The retargeted trajectories are combined with bilateral hand commands mapped from reconstructed hand configurations to form robot action targets. Compared with the original GMR pipeline, our pipeline reduces mean end-effector position error from 16.423 cm to 2.422 cm and mean orientation error from 39.590° to 15.596°, averaged over both end effectors and all valid frames. We further inspect retargeted trajectories for root discontinuities, joint-limit violations, and excessive wrist alignment errors, and exclude low-quality segments.

\noindent\textbf{Corpus curation and dataset splits.}
We curate the full corpus to reduce redundancy and balance scene proportions. We extract V-JEPA~2~\cite{assran2025v} video features and cluster them in a shared representation space. Together with segment-level semantic labels, these clusters guide the removal of redundant demonstrations and adjustment of scene proportions. We then use the same visual and semantic information to select a fixed held-out validation set and construct nested training subsets of increasing scale from the remaining data. Figure~\ref{fig:data_distribution} visualizes the corpus feature distribution through a t-SNE~\cite{maaten2008visualizing} projection, together with representative observations from selected clusters.

\section{\ourMthd{}}
\label{sec:method}

\begin{figure}[!t]
\centering
\captionsetup{type=figure, justification=justified, singlelinecheck=false}
\includegraphics[width=\textwidth]{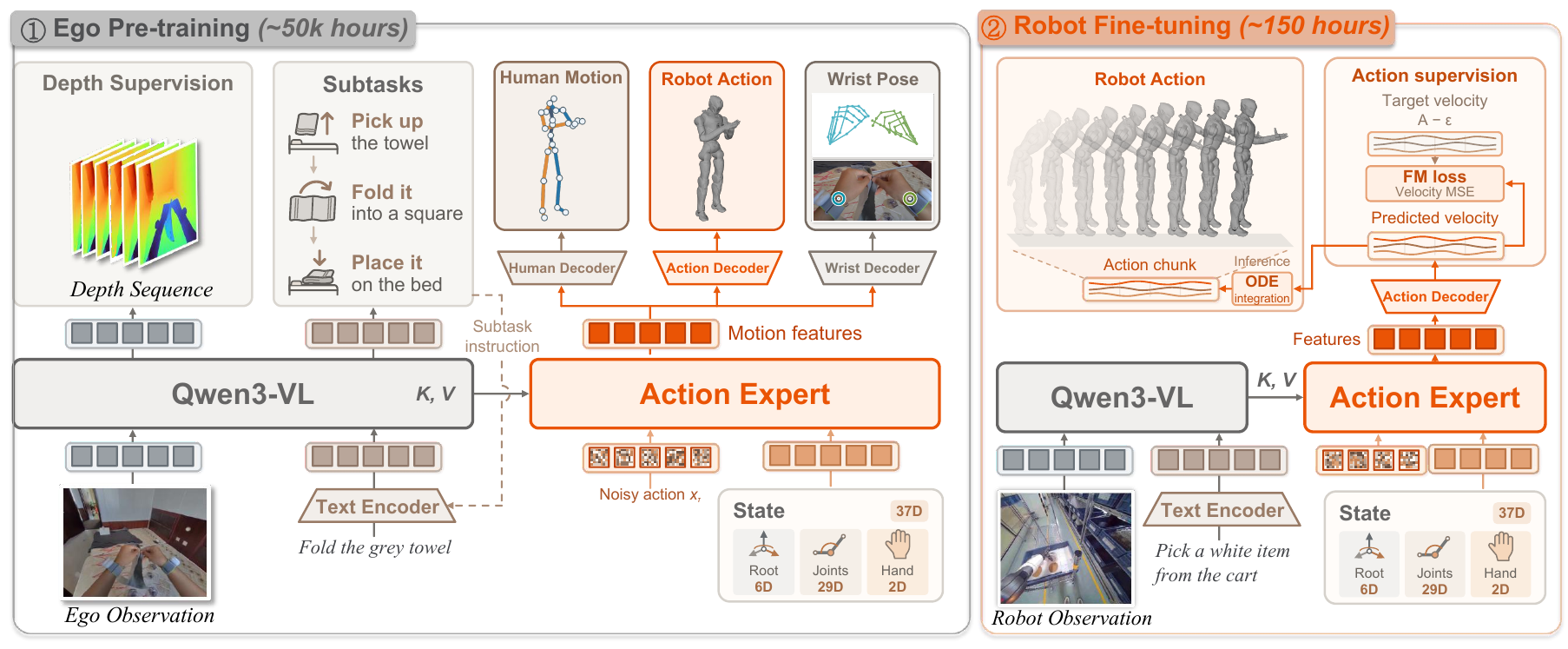}
\caption{\textbf{Two-stage training of \ourMthd{}.} We pretrain \ourMthd{} on \ourData{}, using a shared action expert to jointly predict human motion, wrist poses, and retargeted robot actions, with auxiliary depth and subtask supervision. We then retain the shared expert and robot action branch and fine-tune the policy on real-robot demonstrations.}
\label{fig:method}
\vspace{-6pt}
\end{figure}

An overview of \ourMthd{} is depicted in Figure~\ref{fig:method}.
Through pretraining on whole-body human data and post-training on teleoperation data, it brings human experience to humanoid robots.
Because differences in body proportions can distort retargeted poses, we pretrain on both the original human motion and the retargeted T800 actions to transfer human motion priors into the robot action space.
Additional supervision on wrist poses and scene depth strengthens hand localization and geometric perception.
Post-training then grounds these motion priors in physical robot control.

\subsection{Model Framework}
\label{sec:model_framework}

At time $t$, \ourMthd{} predicts a whole-body action sequence of length $H$ from an RGB image $I_t$ captured by the head camera, a task instruction $\ell_t$, and the current proprioceptive state $\mathbf{s}_t$:
\begin{equation}
\mathbf{A}_t \sim \pi_\theta(\,\cdot\mid I_t,\ell_t,\mathbf{s}_t),\;
\mathbf{A}_t = (\mathbf{a}_{t,k})_{k=0}^{H-1}
\in \mathbb{R}^{H\times 40}.
\label{eq:whole_body_policy}
\end{equation}
During pretraining, the action expert additionally predicts the original human motion and wrist poses over the same horizon.

Qwen3-VL~\cite{bai2025qwen3} serves as the VLM backbone and encodes $I_t$ and $\ell_t$ into token features.
The three branches share the action expert's state encoder and diffusion transformer~\cite{peebles2023scalable}.
The shared DiT attends to the vision-language features through cross-attention.
Each target forms a branch that connects to the action expert through its own input encoder and decoder.
The input encoder embeds a noisy target sequence into a common token space, and the decoder maps the DiT outputs to a velocity in the target space.
The three branches use the same vision-language features and 37-dimensional proprioceptive state, together with their respective flow times, to predict velocity fields in their own target spaces.

Sharing the action expert allows the original human motion to supervise the same parameters used for robot action prediction.
The VLM also supplies image features to the depth decoder and uses its language head to predict the current subtask.
The target representations are defined in Section~\ref{sec:targets}, and the prediction objectives are described in Section~\ref{sec:pretraining}.

\subsection{Action and Motion Representations}
\label{sec:targets}

\textbf{Robot actions.}
Each 40-dimensional action $\mathbf{a}_{t,k}$ specifies a target for T800's low-level controller, consisting of a root displacement $\mathbf{d}_{t,k}\in\mathbb{R}^{3}$, a 6D root rotation $\mathbf{r}_{t,k}\in\mathbb{R}^{6}$, absolute angle targets $\mathbf{q}^{a}_{t+k}\in\mathbb{R}^{29}$ for 29 joints, and bilateral gripper commands $\mathbf{g}^{a}_{t+k}\in\mathbb{R}^{2}$.
These components form $\mathbf{a}_{t,k}=[\mathbf{d}_{t,k};\mathbf{r}_{t,k};\mathbf{q}^{a}_{t+k};\mathbf{g}^{a}_{t+k}]$, where the superscript $a$ denotes action targets.
Root targets are expressed in a local frame fixed at prediction time $t$, with its origin at the current root position $\mathbf{p}_t$, its vertical axis aligned with gravity, and its forward axis along the horizontal projection of the root's forward direction.
With $\mathbf{B}_t$ denoting the $3\times 3$ rotation matrix mapping this frame to world coordinates,
\begin{equation}
\mathbf{d}_{t,k} = \mathbf{B}_t^{\mathsf T}\left(\mathbf{p}^{a}_{t+k}-\mathbf{p}_t\right), \ \mathbf{r}_{t,k} = \rho\!\left(\mathbf{B}_t^{\mathsf T}\mathbf{R}^{a}_{t+k}\right),
\label{eq:heading_frame_action}
\end{equation}
where $\mathbf{p}^{a}_{t+k}$ and $\mathbf{R}^{a}_{t+k}$ are the target root position and orientation, and $\rho(\cdot)$ concatenates the first two columns of a rotation matrix into a six-dimensional representation~\cite{zhou2019continuity}.
Expressing root targets in this frame makes them independent of the world origin and absolute heading.
Root displacement and rotation at every step of the chunk are expressed in the same frame fixed at time $t$.
The 37-dimensional state $\mathbf{s}_t$ comprises the 6D root orientation in the same frame, 29 joint angles, and two gripper states.
During pretraining, both states and actions come from retargeting the recorded human motion to T800.

\textbf{Human motion.}
Retargeted actions place each demonstration in T800's action space, but morphological differences introduce global errors and distorted postures.
By directly learning the original human motion, the model acquires motion priors unaltered by retargeting.
Starting from the 541-dimensional skeletal representation that incorporates visually reconstructed hand motion, we construct a 161-dimensional human motion target covering the body and both hands.
The body component comprises the headset's 3D position and 6D rotation, the positions of 24 body keypoints, and the 6D rotations of the pelvis, elbows, and wrists.
The hand component comprises eight keypoints per hand, namely the palm center, the index and little-finger metacarpals, and the five fingertips, together with a continuous closure value for each hand.
Headset and body poses are expressed in a gravity-aligned heading frame anchored at the pelvis at the first step of the chunk, and hand keypoints in their respective wrist-local frames.
Positions are normalized with quantile statistics, while 6D rotations and closure values keep their original scales.
The chunk length and frame rate match those of the robot actions.

\textbf{Wrist poses.}
To improve the accuracy of hand-dependent manipulation such as grasping and placement~\cite{zhao2026halomi}, we additionally predict human wrist poses. This 18-dimensional target comprises bilateral wrist positions and 6D rotations in the current camera frame.

\subsection{Pretraining Objectives}
\label{sec:pretraining}

\textbf{Motion prediction.}
We use flow matching~\cite{lipman2022flow} for all three motion objectives.
For branch $b\in\{\mathrm{act},\mathrm{human},\mathrm{wrist}\}$, let $\mathbf{X}^{b}$ denote its target sequence, $M^{b}$ its validity mask, and $\mathbf{c}_t$ the VLM features together with the proprioceptive state $\mathbf{s}_t$.
We sample Gaussian noise $\boldsymbol{\epsilon}\sim\mathcal{N}(\mathbf{0},\mathbf{I})$ and a flow time $\tau_b\in[0,1]$, and minimize
\begin{equation}
\begin{gathered}[c]
\mathbf{X}_{\tau_b}^{b}
= (1-\tau_b)\boldsymbol{\epsilon}+\tau_b\mathbf{X}^{b},\\
\mathcal{L}_{b}
= \mathbb{E}\!\left[\mathcal{R}_{b}\!\left(
v_{\theta}^{b}(\mathbf{X}_{\tau_b}^{b},\tau_b\mid\mathbf{c}_t)
-(\mathbf{X}^{b}-\boldsymbol{\epsilon}),\, M^{b}\right)\right],
\end{gathered}
\label{eq:motion_flow_matching}
\end{equation}
where $v_{\theta}^{b}$ is the predicted velocity and $\mathcal{R}_{b}$ aggregates squared velocity errors over valid target elements.
For the T800 action and wrist branches, we average over all valid elements. For human motion, the mask excludes unreliable channels based on tracking status, motion speed, and skeletal geometric consistency. It also excludes time steps that extend beyond the end of the recorded episode.
We partition the 161-dimensional human target into 14 groups by body part and attribute, average squared errors over valid elements within each group, and take a weighted average over the nonempty groups.

\textbf{Scene depth.}
We supervise current and future depth to strengthen geometric perception. A DPT~\cite{ranftl2021vision} decoder predicts current depth from a fixed-resolution 2D grid of VLM image features.
A second decoder uses the image features, proprioceptive state, and each future time offset to predict a depth residual. We add each residual to a stop-gradient copy of the current prediction.
For both current and future depth, we supervise valid pixels against stereo-derived targets in normalized log-depth space using a SiLog-style term~\cite{eigen2014depth} and a SmoothL1 penalty.

\textbf{Subtask prediction.}
Following $\pi_{0.5}$~\cite{pi0.5}, the VLM language head predicts a description of the current subtask from the head-camera image $I_t$ and the overall task instruction $\ell_t$.
We compute the cross-entropy loss $\mathcal{L}_{\mathrm{subtask}}$ over the valid tokens of each segment's annotated subtask description.

\subsection{Training Procedure}
\label{sec:posttraining}

\textbf{Egocentric pretraining.}
We jointly pretrain the vision-language backbone, shared action expert, and prediction heads on \ourData{}.
The T800 action, human motion, and wrist branches make separate forward passes through the shared action expert.
Each branch trains its own input encoder and decoder, while the shared expert receives gradients from all three motion losses.
We compute each loss over valid target annotations and minimize the weighted sum
\begin{equation}
\begin{aligned}
\mathcal{L}={}&\mathcal{L}_{\mathrm{act}}+\lambda_{\mathrm{h}}\mathcal{L}_{\mathrm{human}}+\lambda_{\mathrm{w}}\mathcal{L}_{\mathrm{wrist}}\\
&+\lambda_{\mathrm{c}}\mathcal{L}_{\mathrm{depth}}^{\mathrm{cur}}+\lambda_{\mathrm{f}}\mathcal{L}_{\mathrm{depth}}^{\mathrm{fut}}+\lambda_{\mathrm{s}}\mathcal{L}_{\mathrm{subtask}}.
\end{aligned}
\label{eq:overall_learning_objective}
\end{equation}

\textbf{Robot post-training.}
We initialize the robot policy with the pretrained vision-language backbone, shared action expert, and T800-specific action encoder and decoder.
The teleoperation demonstrations provide measured robot states and recorded control targets in the same T800 representation used for pretraining.
We fine-tune all parameters of these modules on 150 hours of industrial and household teleoperation data using only the robot action flow-matching loss $\mathcal{L}_{\mathrm{act}}$.

\textbf{Deployment.}
At deployment, we integrate the robot action branch's velocity field from noise and inverse-normalize the result to obtain the action sequence $\mathbf{A}_t$.
The policy regenerates this sequence from the latest observations in a receding-horizon loop~\cite{chi2025diffusion}, converts the root targets to the controller's pose representation, and passes the resulting action targets to T800's low-level controller for execution.

\section{Experiments}
\label{sec:experiments}

We first study how scaling whole-body egocentric pretraining data affects action prediction and downstream task performance.
We then compare \ourMthd{} pretrained on the full corpus with VLA baselines on whole-body loco-manipulation tasks and evaluate its generalization.
Finally, we ablate human motion, wrist pose, and depth supervision to assess their contributions.

\begin{figure}[!t]
\centering
\captionsetup{justification=justified,singlelinecheck=false}
\includegraphics[width=\linewidth]{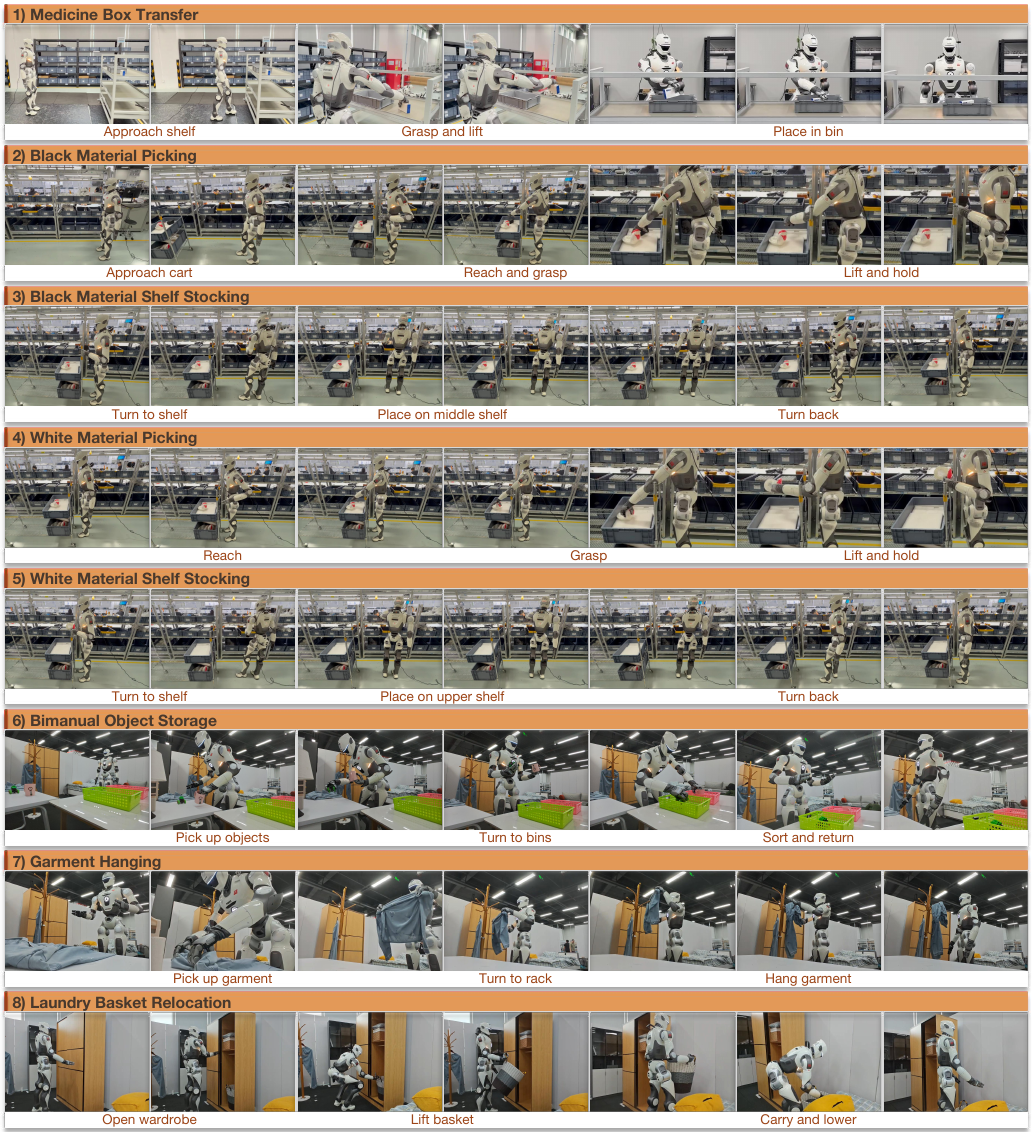}
\caption{\textbf{Eight whole-body loco-manipulation tasks on the EngineAI T800.} Each row shows the key stages of one task from left to right. For long-horizon evaluation, we chain the tasks in rows 2--5 to form the industrial stocking task and those in rows 6--8 to form the household tidying task.}
\label{fig:posttraining_tasks}
\end{figure}

\textbf{Tasks.}
We evaluate eight whole-body loco-manipulation tasks illustrated in Figure~\ref{fig:posttraining_tasks}.
Medicine Box Transfer requires the robot to walk to a shelf, pick up a medicine box, and place it in the bin.
Black Material Picking requires the robot to walk to a cart and pick up the black material.
Black Material Shelf Stocking requires the robot to turn while holding the black material, place it on the middle shelf, withdraw its hand, and turn back.
White Material Picking requires the robot to walk to a cart and pick up the white material.
White Material Shelf Stocking requires the robot to turn while holding the white material, place it on the upper shelf, withdraw its hand, and turn back.

Bimanual Object Storage requires the robot to walk to a table and pick up a cup and a toy car with one hand each.
The robot then turns to the storage table and places the cup in the pink bin and the toy car in the green bin.
Garment Hanging requires the robot to walk to a table, pick up a garment, and turn toward a rack to hang it.
Laundry Basket Relocation requires the robot to walk to a wardrobe, open its door, and crouch to lift a laundry basket.
The robot then steps backward and carries the basket with both hands before placing it on the floor to the right of the wardrobe.

For long-horizon evaluation, we chain Black Material Picking, Black Material Shelf Stocking, White Material Picking, and White Material Shelf Stocking into an industrial stocking task.
Bimanual Object Storage, Garment Hanging, and Laundry Basket Relocation form the household tidying task, in that order.

\textbf{Experimental setup.}
We conduct experiments on EngineAI T800 using approximately 150 hours of teleoperation demonstrations for robot post-training.
Each trajectory synchronizes camera observations, measured robot states, and control targets, with language instructions.
Our baselines are $\pi_{0.5}$~\cite{pi0.5} and GR00T N1.7~\cite{gr00t}.
Each method uses one multitask policy for all eight tasks, post-trained on this dataset for 30,000 steps with a global batch size of 256.
For the main comparison, we initialize \ourMthd{} from the checkpoint pretrained on \ourData{}.
The baselines otherwise follow their officially recommended configurations, and all models run inference on a single NVIDIA RTX 5090 GPU.

\textbf{Evaluation protocol.}
We evaluate each method in ten trials per individual task and per long-horizon task.
For each individual or long-horizon task, methods share ten predefined initial configurations specifying robot poses, object states, and target regions, with variations in starting position and object placement.
Standalone evaluations of Black Material Shelf Stocking and White Material Shelf Stocking begin with the corresponding material already held.
An individual trial succeeds when the robot autonomously completes the task's core manipulation goals.
In each long-horizon trial, the robot and scene are not reset between component tasks, and success requires consecutive completion of all component tasks.
The main comparison covers all eight individual tasks and both long-horizon tasks; real-robot evaluations for scaling and auxiliary-supervision ablations are restricted to the three household tasks.

\subsection{Scaling Human Experience}
\label{sec:ego_pretraining_scaling}

\begin{figure}[!t]
\centering
\captionsetup{justification=justified,singlelinecheck=false}
\includegraphics[width=\linewidth]{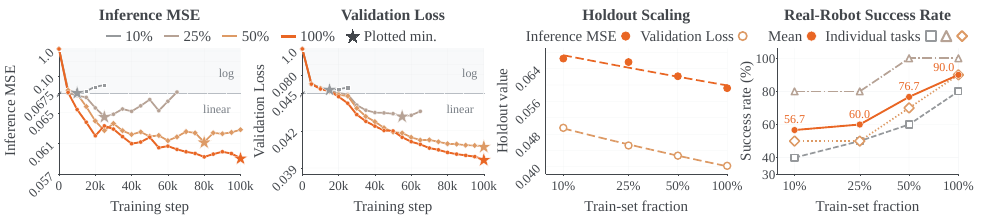}
\caption{\textbf{Scaling whole-body egocentric pretraining improves action prediction and real-robot success.} From left to right, the panels show inference MSE, validation loss, empirical power-law fits for both metrics, and real-robot success rates after post-training. Stars mark the lowest plotted value on each curve in the first two panels. In the final panel, \textcolor[HTML]{96999C}{\(\square\)}, \textcolor[HTML]{B6A295}{\(\triangle\)}, and \textcolor[HTML]{DD9A65}{\(\Diamond\)} denote Bimanual Object Storage, Garment Hanging, and Laundry Basket Relocation, respectively. The solid orange line shows their mean success rate, which rises from 56.7\% to 90.0\% as pretraining data increases from 5,000 to 50,000 hours.}
\label{fig:data_scaling}
\end{figure}

We pretrain \ourMthd{} on four nested training subsets of \ourData{} containing 5,000, 12,500, 25,000, and 50,000 hours of whole-body egocentric demonstrations, corresponding to 10\%, 25\%, 50\%, and 100\% of the corpus.
The runs share the same initialization, optimizer, and learning-rate settings, with a global batch size of 4,096 and a maximum budget of 100,000 updates.
Training is stopped early when inference MSE continues to rise while validation loss plateaus.

We evaluate all scales on the same 100-hour held-out set.
Validation loss measures the flow-matching objective of the robot action branch, while inference MSE measures the discrepancy between generated T800 actions and their targets in normalized action space.
To describe data exposure, we define equivalent epochs as $E=sB/N_{\mathrm{train}}$, where $s$ is the update count, $B$ is the global batch size, and $N_{\mathrm{train}}$ is the number of training frames at each scale.
For offline scaling fits, we use checkpoints matched by equivalent epoch and report both metrics from each checkpoint.

\textbf{Empirical scaling law.}
We fit inference MSE and validation loss separately with the empirical power law $L(D)=AD^{-\alpha}$~\cite{lin2025data}.
Here, $D$ is the fraction of the full training corpus and $A$ is a fitted coefficient.
The fitted exponents are 0.051 for inference MSE and 0.090 for validation loss, as shown in Figure~\ref{fig:data_scaling}.
Increasing pretraining data from 5,000 to 50,000 hours reduces inference MSE from 0.06649 to 0.05926, an improvement of approximately 11\%.
Validation loss decreases from 0.04955 to 0.04025, a reduction of 18.8\%.

\textbf{Downstream task performance.}
We next evaluate whether the observed scaling gains translate to downstream task performance.
At each scale, we initialize the policy from the checkpoint with the lowest held-out inference MSE.
After robot post-training, we evaluate each policy on the three household tasks with ten trials per task.
As shown in Figure~\ref{fig:data_scaling}, scaling human pretraining data from 5,000 to 50,000 hours raises mean success from 56.7\% to 90.0\%.

These results support an empirical scaling law for action prediction in whole-body human pretraining.
They further show that scaling gains translate into improved downstream task performance after robot post-training.

\subsection{Real-Robot Task Performance}
\label{sec:whole_body_task_performance}

We compare \ourMthd{} with existing baselines on EngineAI T800 to evaluate its ability to coordinate locomotion and manipulation across individual tasks and long-horizon tasks.

\textbf{Results.}
\ourMthd{} leads on all eight tasks in Table~\ref{tab:whole_body_success}, with a mean success rate of 93.75\% that exceeds those of $\pi_{0.5}$ and GR00T N1.7 by 36.25 and 35.0 percentage points, respectively.
In particular, it achieves 100\% success on Black Material Picking, Black Material Shelf Stocking, and White Material Shelf Stocking.

Across trials with different starting positions and object placements, we observe \ourMthd{} adjusting its stance for grasping after walking and its hand position for placement at different shelf heights after turning.
The baseline policies, however, are more sensitive to these variations.

The household tasks further require the robot to coordinate bimanual manipulation with whole-body movements such as bending and crouching during grasping and placement.
\ourMthd{} performs well on these tasks, with 80\% success on Bimanual Object Storage and 90\% on Laundry Basket Relocation.

During consecutive execution, \ourMthd{} achieves 90\% success on the industrial stocking task and 80\% on the household tidying task.
The corresponding success rates are 10\% and 20\% for $\pi_{0.5}$, and 30\% and 10\% for GR00T N1.7.
These long-horizon results reflect the model's ability to adapt across stages, as the stance and body orientation left by one stage affect the execution of the next.
\ourMthd{} makes small postural and motion adjustments to accommodate these changing body configurations and complete subsequent stages.

\begin{table}[!t]
\centering
\small
\setlength{\tabcolsep}{5pt}
\renewcommand{\arraystretch}{1.1}
\caption{\textbf{Real-robot task success on the EngineAI T800.} We evaluate each method in ten trials per task and report success rates as percentages. The mean is computed over the eight individual tasks. Long-horizon success requires completing all stages in order without resetting the robot or the scene.}
\label{tab:whole_body_success}

\begin{tabular}{@{}lccc@{}}
\toprule
Task & $\pi_{0.5}$ & GR00T N1.7 & \ourMthd{} \\
\midrule
\multicolumn{4}{@{}l}{\textit{Individual tasks}} \\
Medicine Box Transfer & 60 & 70 & \textbf{90} \\
Black Material Picking & 60 & 40 & \textbf{100} \\
Black Material Shelf Stocking & 50 & 70 & \textbf{100} \\
White Material Picking & 60 & 60 & \textbf{90} \\
White Material Shelf Stocking & 60 & 80 & \textbf{100} \\
Bimanual Object Storage & 40 & 30 & \textbf{80} \\
Garment Hanging & 80 & 80 & \textbf{100} \\
Laundry Basket Relocation & 50 & 40 & \textbf{90} \\
\midrule
Mean (individual tasks) & 57.5 & 58.75 & \textbf{93.75} \\
\midrule
\multicolumn{4}{@{}l}{\textit{Long-horizon tasks}} \\
Industrial stocking task & 10 & 30 & \textbf{90} \\
Household tidying task & 20 & 10 & \textbf{80} \\
\bottomrule
\end{tabular}
\end{table}

\begin{table}[!t]
\centering
\small
\setlength{\tabcolsep}{5pt}
\renewcommand{\arraystretch}{1.15}
\caption{\textbf{Generalization to new instruction combinations and unseen objects.} For each of four new pairings of familiar objects and destinations, we report the placement success rate over ten trials per method. Tissues and a medicine bottle, both unseen during robot post-training, appear together in ten additional trials per method. We report a separate completion rate for each object.}
\label{tab:generalization}
\begin{tabularx}{\linewidth}{@{}l*{6}{>{\centering\arraybackslash}X}@{}}
\toprule
& \multicolumn{4}{c}{Instruction combinations}
& \multicolumn{2}{c}{Unseen objects} \\
& \multicolumn{4}{c}{Task success (\%) $\uparrow$}
& \multicolumn{2}{c}{Object-level completion (\%) $\uparrow$} \\
\cmidrule(lr){2-5}\cmidrule(lr){6-7}
Method
& \makecell{Black material\\$\rightarrow$ upper shelf}
& \makecell{White material\\$\rightarrow$ middle shelf}
& \makecell{Toy car\\$\rightarrow$ pink bin}
& \makecell{Cup\\$\rightarrow$ green bin}
& Tissues
& \makecell{Medicine\\bottle} \\
\midrule
$\pi_{0.5}$ & 20 & 0 & 10 & 10 & 0 & 0 \\
GR00T N1.7 & 40 & 0 & 20 & 10 & 0 & 0 \\
\midrule
\textbf{\ourMthd{}} & \textbf{100} & \textbf{100} & \textbf{80} & \textbf{80} & \textbf{50} & \textbf{30} \\
\bottomrule
\end{tabularx}
\end{table}

\subsection{Generalization}
\label{sec:generalization}

We further evaluate whether the models generalize to instruction combinations and objects absent from the robot post-training demonstrations.
All evaluations use the corresponding post-trained models from Section~\ref{sec:whole_body_task_performance} without additional training.

\textbf{Compositional generalization.}
We form four new test pairings using objects and destinations already present in the robot post-training demonstrations.
In the industrial tasks, the robot is instructed to place black material on the upper shelf and white material on the middle shelf, reversing the destinations used in post-training.
In the household tasks, the robot is similarly instructed to place the toy car in the pink bin and the cup in the green bin.
We evaluate each pairing separately in ten trials per method, with success defined as placing the correct object at the destination specified by the instruction.

\ourMthd{} outperforms both baselines on all four instruction tests in Table~\ref{tab:generalization}, with 100\% success on both industrial tasks and 80\% on both household tasks.
We observe the baseline models placing objects at their original training destinations or stopping between the original and newly instructed destinations, indicating that the original targets continue to influence execution after the instruction changes.
By contrast, \ourMthd{} adjusts the placement goal according to the new instruction and reuses the learned manipulation behavior to complete placement at the specified destination.

\textbf{Generalization to unseen objects.}
We replace the two objects in Bimanual Object Storage with tissues and a medicine bottle, both absent from the robot post-training demonstrations.
We evaluate each method in ten trials with both replacement objects present and score completion separately for each object.

\ourMthd{} achieves object-level completion rates of 50\% for tissues and 30\% for the medicine bottle, while both baselines achieve 0\% for both objects, as reported in Table~\ref{tab:generalization}.
\ourMthd{} shows better generalization to objects absent from robot post-training.

\subsection{Ablation Studies}
\label{sec:ablation}

\begin{table}
\centering
\small
\setlength{\tabcolsep}{2.5pt}
\renewcommand{\arraystretch}{1.15}
\caption{\textbf{Ablation of human motion, wrist pose, and depth supervision.} We pretrain all variants on the same 5,000 hours of \ourData{} data. Before robot post-training, we report MSE on normalized actions, mean 3D Euclidean position errors for the root and each wrist, and joint-angle MAE for the whole body and legs. After post-training, we report success rates for three household tasks and their mean, with ten real-robot trials per task for each variant.}
\label{tab:auxiliary_ablation}
\begin{tabular*}{\linewidth}{@{\extracolsep{\fill}}lrrrrrrcccc@{}}
\toprule
& \multicolumn{6}{c}{Offline prediction error $\downarrow$}
& \multicolumn{4}{c}{Real-robot success (\%) $\uparrow$} \\
\cmidrule(lr){2-7}\cmidrule(lr){8-11}
Configuration
& \makecell{Action\\MSE}
& \makecell{Root\\(cm)}
& \makecell{Left\\wrist\\(cm)}
& \makecell{Right\\wrist\\(cm)}
& \makecell{Whole body\\MAE ($^\circ$)}
& \makecell{Legs\\MAE ($^\circ$)}
& \makecell{Bimanual\\Object\\Storage}
& \makecell{Garment\\Hanging}
& \makecell{Laundry\\Basket\\Relocation}
& Mean \\
\midrule
Base & 0.092785 & 6.48 & 9.31 & 10.09 & 7.88 & 3.73 & 0 & 40 & 0 & 13.3 \\
+ Wrist & 0.092068 & 6.43 & 9.30 & 10.04 & 7.91 & 3.73 & 20 & 30 & 10 & 20.0 \\
+ Human Motion & 0.089660 & 6.35 & 9.12 & 9.83 & 7.72 & 3.58 & \textbf{30} & 60 & 40 & 43.3 \\
+ Depth & 0.091593 & 6.42 & 9.30 & 9.89 & 7.81 & 3.69 & 20 & 70 & 30 & 40.0 \\
\midrule
\textbf{Full} & \textbf{0.088804} & \textbf{6.23} & \textbf{9.04} & \textbf{9.81} & \textbf{7.65} & \textbf{3.47} & \textbf{30} & \textbf{80} & \textbf{50} & \textbf{53.3} \\
\bottomrule
\end{tabular*}
\end{table}

\begin{figure}[!t]
\centering
\captionsetup{justification=justified,singlelinecheck=false}
\includegraphics[width=0.95\linewidth]{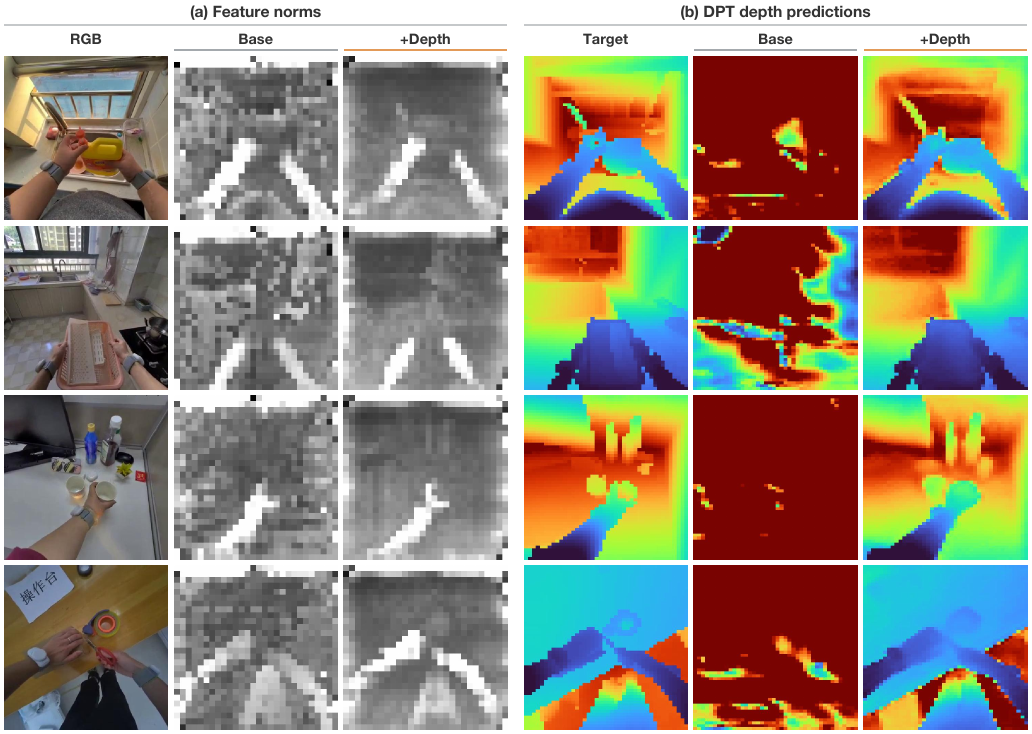}
\caption{\textbf{Effect of auxiliary depth supervision on visual representations.} The left panel shows the RGB inputs and the \(\ell_2\)-norm maps of the final-layer image features for both models. We train a separate DPT probe on each frozen backbone under identical settings. The model trained with depth supervision better recovers the relative depths of the arms, supporting surfaces, and background shown in the stereo-derived targets.}
\label{fig:depth_analysis}
\end{figure}

We conduct controlled ablations to examine the contributions of human whole-body and hand motion, wrist pose, and depth supervision during pretraining.
We use a separate pretraining configuration for all ablations: 5,000 hours of whole-body egocentric data, 50,000 steps, and 64 GPUs with a batch size of 16 per GPU, followed by the same robot post-training.
Base uses only retargeted T800 action supervision; each single-objective variant adds one auxiliary objective, and Full combines all three.

On the validation set used in Section~\ref{sec:ego_pretraining_scaling}, we report MSE for normalized 40-dimensional actions, mean 3D Euclidean root and wrist position errors, and joint-angle MAE for the 29 body joints and 12 leg joints, with wrist positions computed relative to the body using T800 forward kinematics.
For real-robot evaluation, we test each configuration on Bimanual Object Storage, Garment Hanging, and Laundry Basket Relocation, with ten trials per task.

Full achieves the lowest errors on all six offline metrics in Table~\ref{tab:auxiliary_ablation}, reducing action MSE by 4.29\% relative to Base.
Among the single-objective variants, Human Motion performs best on all six offline metrics, including root and wrist position errors and body and leg joint-angle errors.
Depth also reduces all six errors relative to Base, whereas Wrist only slightly reduces wrist position errors and leaves leg-joint MAE unchanged.

After post-training, Full achieves the highest mean real-robot success rate in Table~\ref{tab:auxiliary_ablation}, at 53.3\% compared with 13.3\% for Base.
Among the single-objective variants, Human Motion has the highest mean success rate, while Depth is strongest on Garment Hanging, with 70\% success.
Wrist supervision does not improve every task: garment-hanging success decreases from 40\% for Base to 30\%.
Full's mean real-robot success rate also exceeds that of the strongest single-objective variant by 10.0 percentage points: combining the objectives yields additional gains in real-robot execution.

We compare the visual representations of Base and Base + Depth, with a separate DPT depth probe trained on each frozen model under the same settings.
In Figure~\ref{fig:depth_analysis}, the depth-supervised model produces feature-norm maps in which the arms stand out against a smoother background.
Its depth predictions recover the relative depth of foreground arms, supporting surfaces, and background structures, whereas Base produces large, nearly uniform regions.

\section{Conclusion}
\label{sec:conclusion}

We presented \ourMthd{}, which jointly predicts original human motion and retargeted robot actions through a shared action expert.
We constructed \ourData{}, pairing 50,000 hours of egocentric stereo video with synchronized whole-body motion. 
Scaling pretraining data reduces held-out action prediction error in line with an empirical power law and improves downstream task success. 
After post-training, a single policy achieves a mean success rate of 93.75\% across eight individual loco-manipulation tasks. 
Original human motion thus remains an informative pretraining target even after retargeting.
\newpage
\section{Contributions}
\label{sec:contributions}

\begingroup
\setlength{\parindent}{0pt}
\setlength{\parskip}{0.8\baselineskip}

\newcommand{\projectleadmark}{\textsuperscript{\scalebox{1.2}{\textbf{\dag}}}}
Contributors to \ourMthd{} are listed alphabetically by last name within each group.
\projectleadmark\,{denotes the project lead.}

\raggedright
\textbf{Core Contributors.}\quad
Shuiqi~Cheng, Boren~Jiang, Weilong~Ju, Liyun~Li, Xuanyi~Li\projectleadmark,
Chujie~Qin, and Kai~Wang.

\textbf{Contributors.}\quad
Kun~Cheng, Haoyin~Fan, Jiarong~Feng, Jingbo~Gong, Alan~Liu, Shuixin~Luo, Yang~Wang, Yucheng~Zhang, and Zilong~Zhang.

\textbf{Acknowledgments.}\quad
For their time and support throughout the \ourMthd{} project, we thank Pengju~An, Kaiyong~Chen, Zisong~Chen, Yuqi~Ding, Fei~Gao, Bowen~Gong, Qinbo~Hu, Jun~Huang, An~Li, Xinyu~Li, Binbin~Liang, Shuangran~Liang, Bingxi~Liu, Junyu~Mei, Junning~Qiu, Shaobin~Ren, Yi~Ren, Zichao~Shen, Lingjie~Song, Haiyan~Wang, Shunyu~Wang, Xu~Yang, Zhixing~Yuan, Ziyin~Yue, Wancheng~Zhang, Yifan~Zhang, Feiyu~Zhao, and Pengpeng~Zhao.

\par
\endgroup

\clearpage
\setcitestyle{numbers}
\bibliographystyle{plainnat}
\bibliography{main}

\end{document}